\documentclass[runningheads,a4paper]{llncs}

\usepackage{amssymb}
\usepackage{graphicx}
\usepackage[]{algorithm2e}

\usepackage{url}
\urldef{\mailsa}\path|{rrychtarikova, stys, psoucek}@frov.jcu.cz|
\newcommand{\keywords}[1]{\par\addvspace\baselineskip
\noindent\keywordname\enspace\ignorespaces#1}

\begin{document}

\mainmatter  

\title{Colorimetric Calibration of a Digital Camera}

\titlerunning{Calibration of Digital Cameras}

%
%
\author{Renata Rycht\'{a}rikov\'{a} \and Pavel Sou\v{c}ek \and Dalibor \v{S}tys }

\authorrunning{Rycht\'{a}rikov\'{a} et al.} 

\institute{University of South Bohemia in \v{C}esk\'{e} Bud\v{e}jovice, Faculty of Fisheries and Protection of Waters, South Bohemian Research Center of Aquaculture and Biodiversity of Hydrocenoses, Institute of Complex Systems, Z\'{a}mek 136, 373 33 Nov\'{e} Hrady, Czech Republic \\
              Tel.: +420-777 729 581\\
             \mailsa\\ 
\url{http://www.frov.jcu.cz/en/institute-complex-systems}}

\date{Received: date / Accepted: date}

%
%

\toctitle{Calibration of Digital Cameras}
\tocauthor{Rycht\'{a}rikov\'{a} et al.}
\maketitle

\begin{abstract}
In this paper, we introduce a novel -- physico-chemical -- approach for calibration of a digital camera chip. This approach utilizes results of measurement of incident light spectra of calibration films of different levels of gray for construction of calibration curve (number of incident photons vs. image pixel intensity) for each camera pixel. We show spectral characteristics of such corrected digital raw image files (a primary camera signal) and demonstrate their suitability for next image processing and analysis.
\end{abstract}

\section{Introduction}

Digital cameras are powerful tools to capture images for use in image processing. Their current engineering and scientific utilization is extensive and includes, e.g., security applications, aerial mapping or microscopy. These applications are very often aimed to detect movements (dynamics) with the resolution of a few pixels or to image processing. For a full utilization of information brought by a camera, it  is necessary to calibrate the camera chip as best as possible which, e.g., makes the consecutive digital images more mutually comparable, facilitates the detection and identification of depicted objects and enables to use the camera as a colorimeter to monitor changes of chemical composition.

In this article, we describe a method which combines an experimental approach of measurement of spectral properties with a software calibration package for correction of the inhomogeneities in raw image files. We show that analyses of uncalibrated images can lead to wrong conclusions when computer analytical algorithms are applied. In one of the steps of image processing, a map of number of photons reaching each light sensitive element at the camera chip is obtained. This map candidates for an ultimately exchangeable data format which enables a comparison between cameras irrespective to their construction. We recommend this physico-chemical approach to image correction to become standard for scientific image inspection, exchange and any computer analysis.

\section{Materials and methods}

\subsection{Image capture} \label{capture}

The testing image comes from an experiment on fish behaviour in a 5iD viewer~\cite{Stys2015} -- an equipment which is composed of an aquarium surrounded by four mirrors and gives thus 5 views of fish positions. The experiment examined an intention of a fish group to exploit free space from which is separated by an obstacle with a small window which allows only one fish to swim at a time through. The 5iD viewer was placed in a light box which ensured the most uniform illumination.

The testing image was captured by a rapid high-resolution JAI Spark SP-20000-USB camera. The camera was adjusted with exposure 170 ms and gain 100. 
The recorded images were stored in a 16-bit grayscale PNG format (resolution 2048$\times$2560). The delay between shots was 500 ms.

The corrected and uncorrected images in Figs.~\ref{Fig2}b, \ref{Fig3}b, \ref{Fig5}a, and \ref{Fig6}a were visualized using the least information lost (LIL) 12- to 8-bit compression~\cite{Stys2016}.

\subsection{Image correction} \label{correction}

\begin{figure}
\centering
\includegraphics[width= \textwidth]{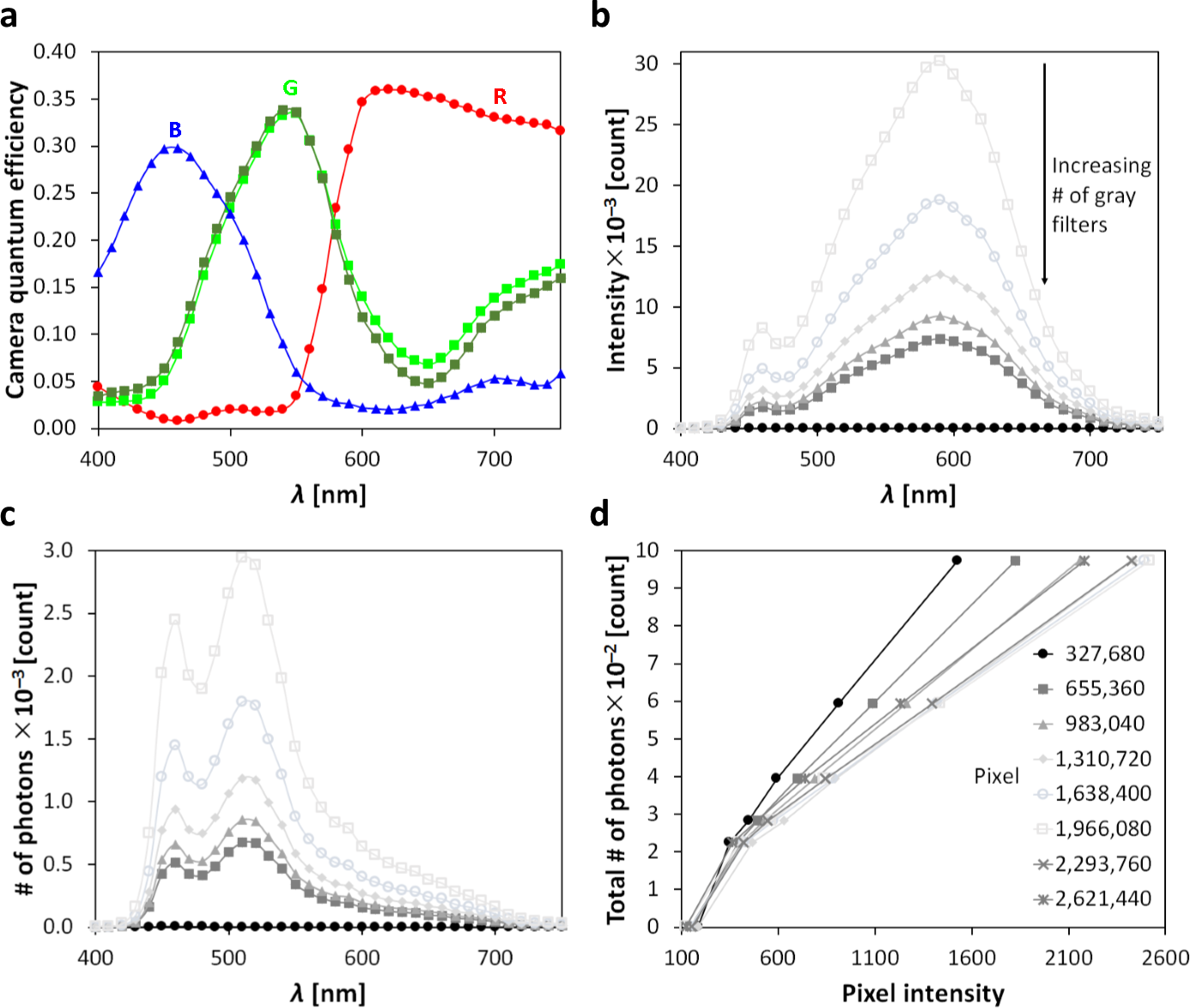}
\caption{(a) Light spectra of grayscale filters measured by a fiber spectrophotometer, (b) declared spectra of RGB camera filters~\cite{JAI}, (c) calculated spectra of incoming light reaching the blue camera channel. Integral under the curve (c) was used as a calibration value for the construction of the calibration curve. (d) Calibration curves for blue camera pixels lying in the same column.}
\label{Fig1}
\end{figure}

The camera calibration and image correction was performed in the following steps:

\begin{enumerate}
\item The camera objective was replaced for a fibre spectrophotometer Ocean Optics USB 4000 VIS-NIR-ES by which the spectra (Fig.~\ref{Fig1}b) of a series of 1--4 gray filters were measured successively. The spectrum of zero and the highest intensity was measured in dark and without any filter, respectively. Images of the sets of filters relevant to the spectra (including the zero and the highest spectrum) were captured in 6 parallels from which a mean calibration image was computed.
\item The light spectra captured by each pixel of the colour camera filter were obtained from multiplication of the measured incident spectra by the respective (red, green, blue) filter profile (supplied by the JAI company~\cite{JAI}; Fig.~\ref{Fig1}a).
\item For each gray filter, total number of photons (i.e., counts) captured by each pixel was calculated as an integral (trapezoidal rule) of the area below the respective incident spectrum in Fig.~\ref{Fig1}c. 
\item For each pixel of the mean calibration image (see item 1), a calibration points were constructed (Fig.~\ref{Fig1}) as a dependency of the pixel intensity on the total number of photons reaching the pixel. Each pair of two consecutive calibration points were fitted by linear interpolation.
\item Using the calibration relation of the relevant section of the calibration curve, the intensity of each pixel of the testing image was converted to values of the total number of photons (in double precision numbers). For further image operations, the resulted matrix was transferred into a 12-bit PNG format.
\end{enumerate}

The preparation of the calibration curve (items 2--4) is the content of Algorithm~\ref{Alg1}. The testing image was then corrected using Algorithm~\ref{Alg2}.

\begin{algorithm} \label{Alg1}
\IncMargin{1em}
\KwIn{\textbf{sQE} as a quantum efficiency spectrum of one colour camera channel;\\ \qquad \quad
\textbf{sFt} as a spectrum of a stack of gray filters ($N$-item folder);\\ \qquad \quad
\textbf{iFt} as a relevant colour channel of the stack of gray filters ($N$-item\\ \qquad \quad folder);}
\KwOut{\textbf{k} as a matrix of the slopes of the linear sections of the calibration curve;\\ \qquad \quad
 \textbf{s} as a matrix of the shifts of the linear sections of the calibration curve;\\ \qquad \quad
 \textbf{int} as a matrix of the intensities which demarcates the ranges of the\\ \qquad \quad linear sections of the calibration curve;}
\BlankLine
\BlankLine

\textbf{A} = zeros(N,1); \qquad \emph{\% create an empty (zero) N-element vector}
\BlankLine
\BlankLine
\For{$i = 1$ \KwTo $N$}{
\textbf{sFt} = readSp($i$);\\
\qquad\emph{\% read a spectrum} \textbf{sFt} \emph{for (i) gray filters}\\
\textbf{wSp} = \textbf{sFlt} .* \textbf{sQE};\\
\qquad\emph{\% for each wavelength, weight the spectrum} \textbf{sFt} \emph{by the spectrum} \textbf{sQE}\\
\textbf{A}($i$) = integrateSpectrum(\textbf{wSp});\\
\qquad\emph{\% integrate the area under the weighted spectrum to obtain a total\\ \qquad \quad number of photons reaching the colour channel of the camera chip}}
\BlankLine
\BlankLine
\textbf{int} = zeros($N,nPx$);\\
\textbf{k} = zeros($N,nPx$);\\
\textbf{s} = zeros($N,nPx$);\\
\qquad \emph{\% create empty (zero) matrices of the output calibration files (i.e., of the\\ \qquad \quad calibration parameters)}
\BlankLine
\BlankLine

\For{$i = 1$ \KwTo $N-1$}{
\textbf{iFt1} = readIm($i$) ;\\
\textbf{iFt2} = readIm($i+1$) ;\\
\qquad\emph{\% read raw image files of relevant colour channel for (i) and (i+1)\\ \qquad \quad gray filters}
\BlankLine
\BlankLine
\For{$j=1$ \KwTo $nPx$}{
\textbf{int}($i,j$) = \textbf{iFt1}($j$); \\
\qquad\emph{\% read and save the first edge point of the section of the calibration\\ \qquad \quad curve}\\
\textbf{k}($i,j$) = (\textbf{A}($i+1$) - \textbf{A}($i$))/(\textbf{iFt2}($j$)- \textbf{iFt1}($j$)); \\
\textbf{s}($i,j$) = \textbf{A}($i$)-\textbf{k}($j$) .* \textbf{iFt1}($j$);\\
\qquad\emph{\% calculate and save a slope and a shift of the relevant section of the\\ \qquad \quad calibration curve}
}
}
\BlankLine
\BlankLine
\caption{Construction of the calibration curve and creation of the calibration file for one colour channel.}
\end{algorithm}

\begin{algorithm} \label{Alg2}
\IncMargin{1em}
\KwIn{\textbf{I} as an original ($nPx$)-resolved uncorrected raw image file;\\ \qquad \quad
\textbf{int}, \textbf{k}, and \textbf{s} as a content of the calibration file (see Algorithm~\ref{Alg1})
}
\KwOut{\textbf{cI} as a corrected image}
\BlankLine
\BlankLine
\textbf{cI} = I .* 0; \qquad\emph{\% create an empty (zero) matrix of the size of the} \textbf{I}
\BlankLine
\BlankLine
\For{$j = 1$ \KwTo $nPx$}{
\uIf{$\textbf{I}(j)  <  \textbf{int}(2,j)$}{
\textbf{cI}($j$) = \textbf{k}(1,$j$) .* \textbf{I}(1,$j$) + \textbf{s}(1,$j$);}
\BlankLine
\BlankLine
\uElseIf{$\textbf{I}(j)  \in  \langle\textbf{int}(2,j),\textbf{int}(3,j))$}{
\textbf{cI}($j$) = \textbf{k}(2,$j$) .* \textbf{I}(2,$j$) + \textbf{s}(2,$j$);}
\BlankLine
\BlankLine
\uElseIf{$\textbf{I}(j)  \in  \langle\textbf{int}(3,j),\textbf{int}(4,j))$}{
\textbf{cI}($j$) = \textbf{k}(3,$j$) .* \textbf{I}(3,$j$) + \textbf{s}(3,$j$);\\ ...}
\BlankLine
\BlankLine
\Else{\textbf{cI}($j$) = \textbf{k}($N$,$j$) .* \textbf{I}($N$,$j$) + \textbf{s}($N$,$j$);}
}
\qquad\emph{\% for intensity of each pixel of the image} \textbf{I}, \emph{find the relevant linear section\\ \qquad \quad of the calibration curve and use its mathematical relation to recalculate \\ \qquad \quad this intensity to the total number of photons}
\BlankLine
\BlankLine
\caption{Image correction of one colour channel.}
\end{algorithm}

\subsection{Thresholding of image intensities}

The corrected and uncorrected testing raw image files were demosaiced into a quarter-resolved image where red and blue pixels of the Bayer mask were adapted to the relevant colour channel and the green pixels were averaged to give the green channel~\cite{Bayer,Tkacik,Ultramicroscopy}. These rgb images were binarized via clustering of the pixels using k-means++ algorithm (squared Euclidian distance)~\cite{kmeans} into two groups (Algorithm~\ref{Alg3}).

\begin{algorithm} \label{Alg3}
\IncMargin{1em}
\LinesNumbered
\KwIn{\textbf{RF} as a ($nPx$)-resolved raw image file;}
\KwOut{\textbf{OutI} as a ($nPx/4$)-resolved binarized image}
\BlankLine
\BlankLine
\textbf{RGB} = demosaic(\textbf{RF});\\
\qquad\emph{\% non-interpolating debayerization of the raw image file~\cite{Tkacik}} \textbf{RF}\\
\textbf{rRGB} = reshape(\textbf{RGB, 3, $nPx$});\\
\qquad\emph{\% reshape the demosaiced} \textbf{RGB} \emph{image into a matrix of the size of 3$\times$nPx,\\ \qquad \quad where columns correspond to the red, green, and blue colour channel,\\ \qquad \quad respectively, and rows correspond to the indices of pixels in a relevant\\ \qquad \quad colour channel}\\
\textbf{noZero} = removeZeros(\textbf{rRGB});\\
\qquad\emph{\% remove columns with zero elements (black pixels of the manually removed\\ \qquad \quad uninformative image background) from the reshaped image} \textbf{rRGB}\\
\textbf{cls} = kmeans(\textbf{noZero,2});\\
\qquad\emph{\% classify colour pixels (columns of the matrix} \textbf{noZero}\emph{) into 2 classes --\\ \qquad \quad class~1 for background, class 2 for signal (objects)}
\BlankLine
\BlankLine
\textbf{OutI} = \textbf{RGB}(:,:,1) .* 0;\\
\qquad\emph{\% create an empty (zero) matrix of the size of one colour channel of the}\\ \qquad \quad \textbf{RGB} \emph{image}\\
$idx$ = 1; \qquad\emph{\% index of the cluster vector} \textbf{cls}
\BlankLine
\BlankLine

\For{j = 1:$(nPx/4)$}{
\If{sum(\textbf{RGB}(j)) $\neq$ 0}{
\qquad \emph{\% if the j$^{th}$ pixel of the image} \textbf{RGB} \emph{is not black...}
\BlankLine
\BlankLine
	\uIf{\textbf{cls}(idx) == 1}{
	\qquad \emph{\% ... and the (idx)$^{th}$ element of the cluster vector} \textbf{cls} {is 1,}\\
	\textbf{OutI}($j$) = 1} \qquad \emph{\% ... label the j$^{th}$ pixel of the output image by 1}\\
	\Else{\textbf{OutI}($j$) = 2\\ \qquad \emph{\% ... else label the j$^{th}$ pixel of the output image by 2}}
\BlankLine
\BlankLine 
$idx$ = $idx$ + 1;
}
}

\BlankLine
\BlankLine
\caption{Bi-level image thresholding of a raw image file.}
\end{algorithm}

\section{Results and discussion}

\subsection{Image correction}

Vignetting is a reduction of brightness or saturation at the periphery of image compared to its center. For scientific and engineering purposes, it is often an unintended and undesired effect caused by camera adjustment or inaccuracy of optics. Such images suffer from strong distortions which disable an exact and simple data processing and analysis.

In order to avoid the vignetting and other image inhomogeneities and fails of camera chip, we propose here a method of calibration (Sect.~\ref{correction}) of intensities of camera pixels which uses an approach typical of spectrophotometry. Light going through a transparent colour material (filter) is gradually absorbed. This absorption of incident radiation occurs in a complementary colour. After the absorption, the remaining radiation reaches the camera Bayer filter, which is an array of red, green, and blue filters which covers a square grid of photosensors. These colour filters identify always only one respective color of visible light and allow the transmitted light to achieve the relevant photosensor (pixel). In CCD (Charge-Coupled Device) or CMOS (Complementary Metal-Oxide-Semiconductor) cameras, the photosensors convert the level of photons into a proportional electrical signal. In this way, the information about the incident light intensity is saved into a digital matrix. Via measurement of light spectra of calibration films of different colours (levels of gray), we shall construct a calibration curve and a new image (data matrix) where, at a given environment (light source and path), each pixel is characterized by a number of incident photons. 

\begin{figure}
\centering
\includegraphics[width=\textwidth]{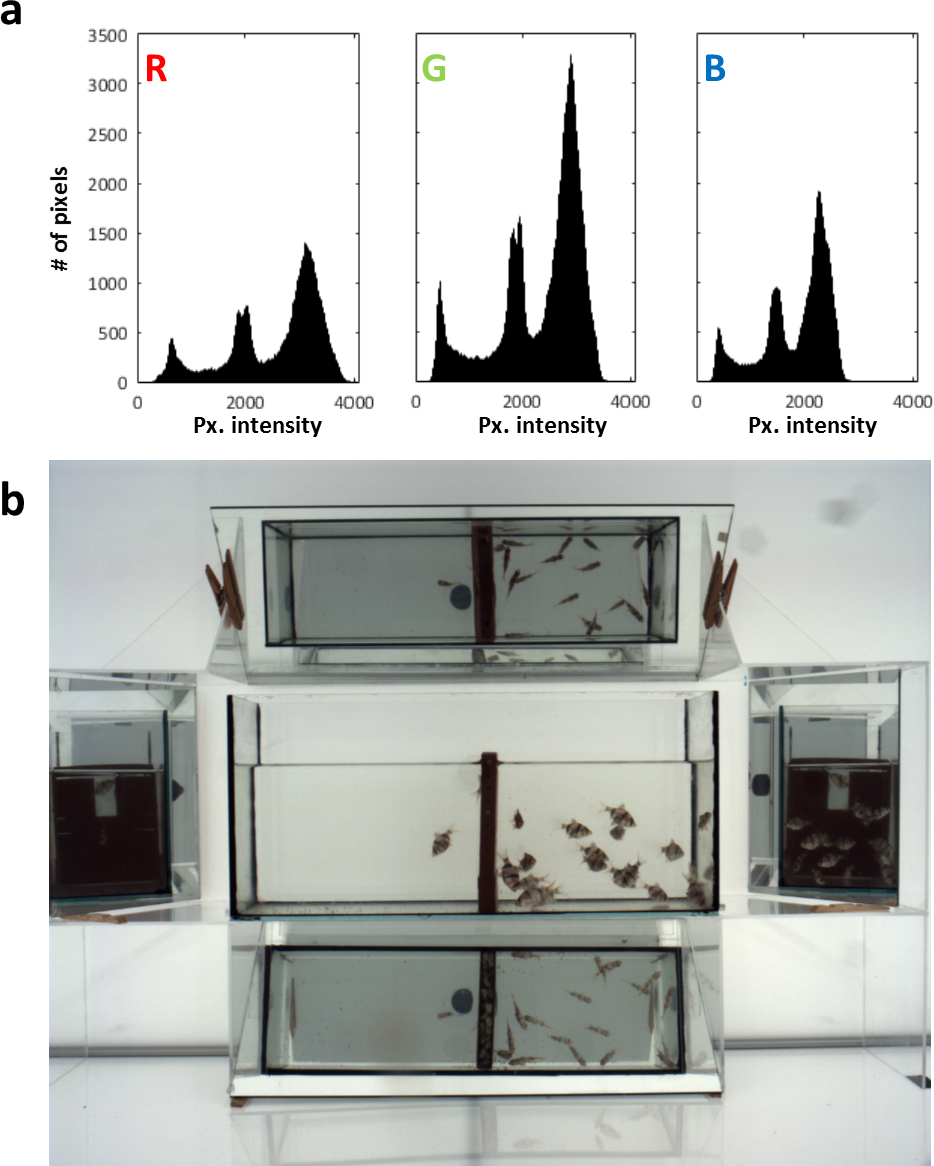}
\caption{Uncorrected image. (a) Intensity histogram in the red (R), green (G), and blue (B) camera channel. (b) 8-bit visualization of the original 12-bit image using the LIL algorithm.}
\label{Fig2}
\end{figure}

\begin{figure}
\centering
\includegraphics[width=\textwidth]{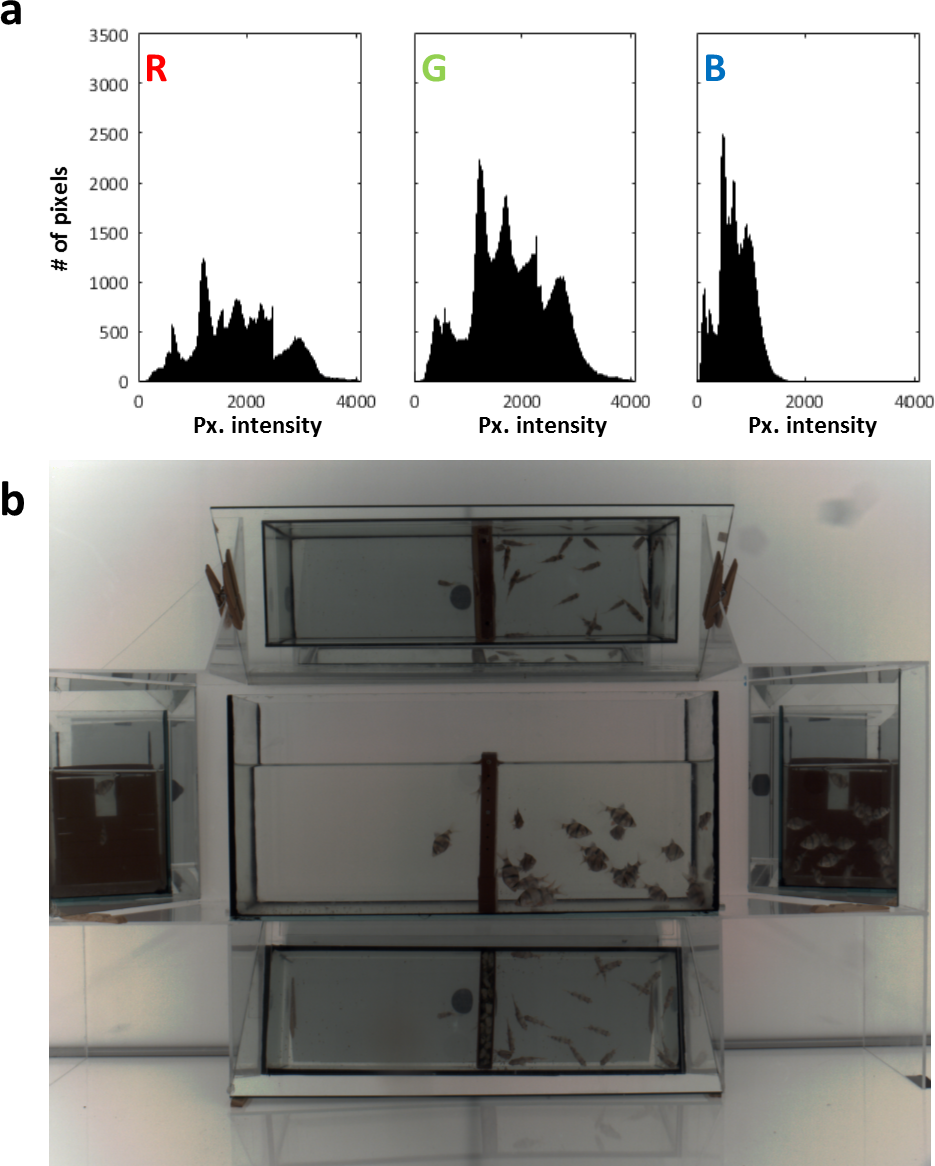}
\caption{Corrected image. (a) Intensity histogram in the red (R), green (G), and blue (B) camera channel. (b) 8-bit visualization of the original 12-bit image using the LIL algorithm.}
\label{Fig3}
\end{figure}

Figs.~\ref{Fig2}--\ref{Fig3} show a result of the image correction for an image from the experiment on exploratory instinct in fish groups (Sect.~\ref{capture}). The camera was set up so that all intensities were safely kept within the intensity range of the camera chip (Fig.~\ref{Fig2}a). The intensity histograms of the corrected image (Fig.~\ref{Fig3}a) differs significantly from the uncorrected image (Fig.~\ref{Fig2}a). The highest intensity peaks (above intensities 2000) in all colour channels in Fig.~\ref{Fig2}a demonstrate that the uncorrected image is dominated by a kind of vignetting which manifests itself in a bright center of Fig.~\ref{Fig2}b). This vignetting was removed after the image correction (Fig.~\ref{Fig3}, where the intensity histograms became more uniform. They correspond to the real-life shadows (Fig.~\ref{Fig3}b).

\subsection{Computer-based image analysis}

Using an image of the fish school, we shall demonstrate here an effect of the described image correction on the detection of the objects of interest. In order to maintain as much original information as possible, we processed a raw discrete signal obtained by the colour camera. 

\begin{figure}
\centering
\includegraphics[width=\textwidth]{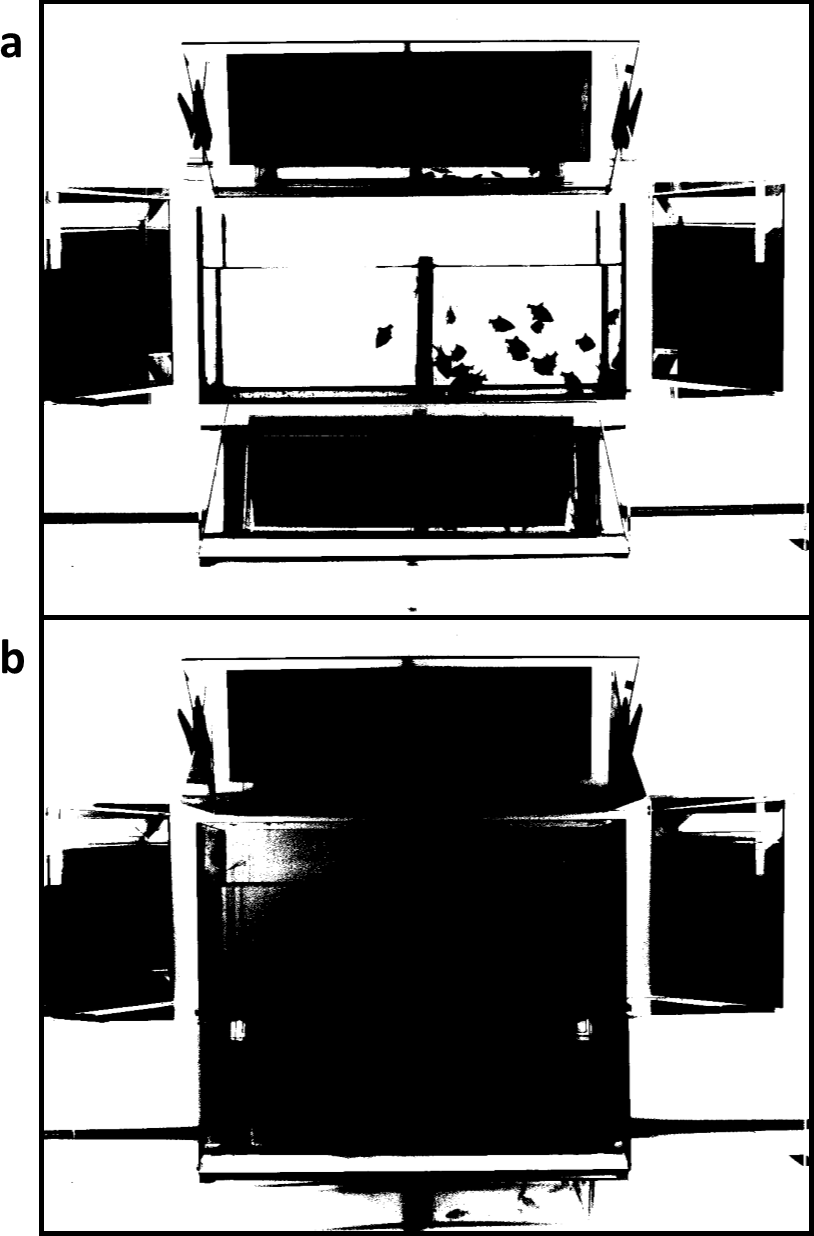}
\caption{Intensity thresholding of the whole uncorrected (a) and corrected (b) image. Colours of pixels were clustered into two groups using $k$-means++ algorithm (squared Euclidian distance).}
\label{Fig4}
\end{figure}

Figs.~\ref{Fig4} shows a result of a simple image thresholding (Algorithm~\ref{Alg3}) via classification of pixels (quadruplets of the Bayer mask) by $k$-means algorithm into two groups -- signal (objects of interest) and background.

\begin{figure}
\centering
\includegraphics[width=\textwidth]{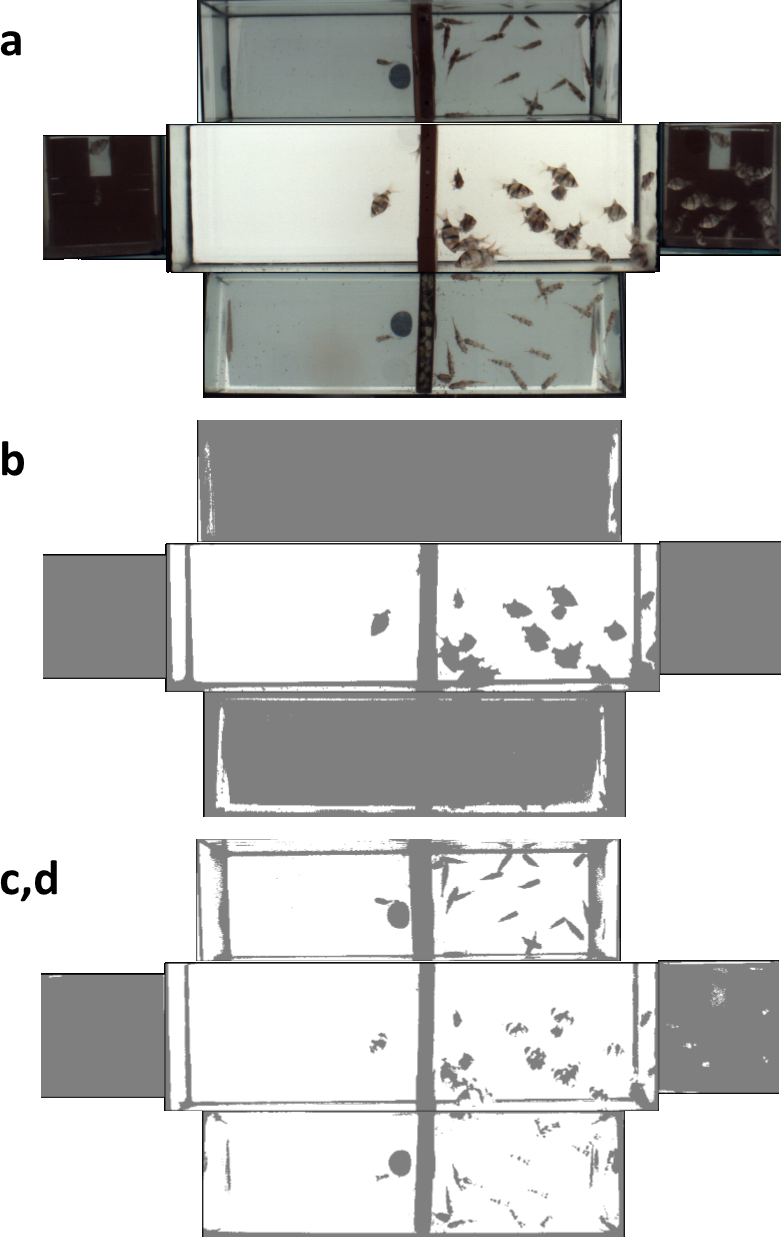}
\caption{Intensity thresholding of the mirrors in the uncorrected image. Colours of pixels were clustered into two groups using $k$-means++ algorithm (squared Euclidian distance). Figures (c) and (d) are mutually inverted. The uninformative black pixels were removed.}
\label{Fig5}
\end{figure}

\begin{figure}
\centering
\includegraphics[width=\textwidth]{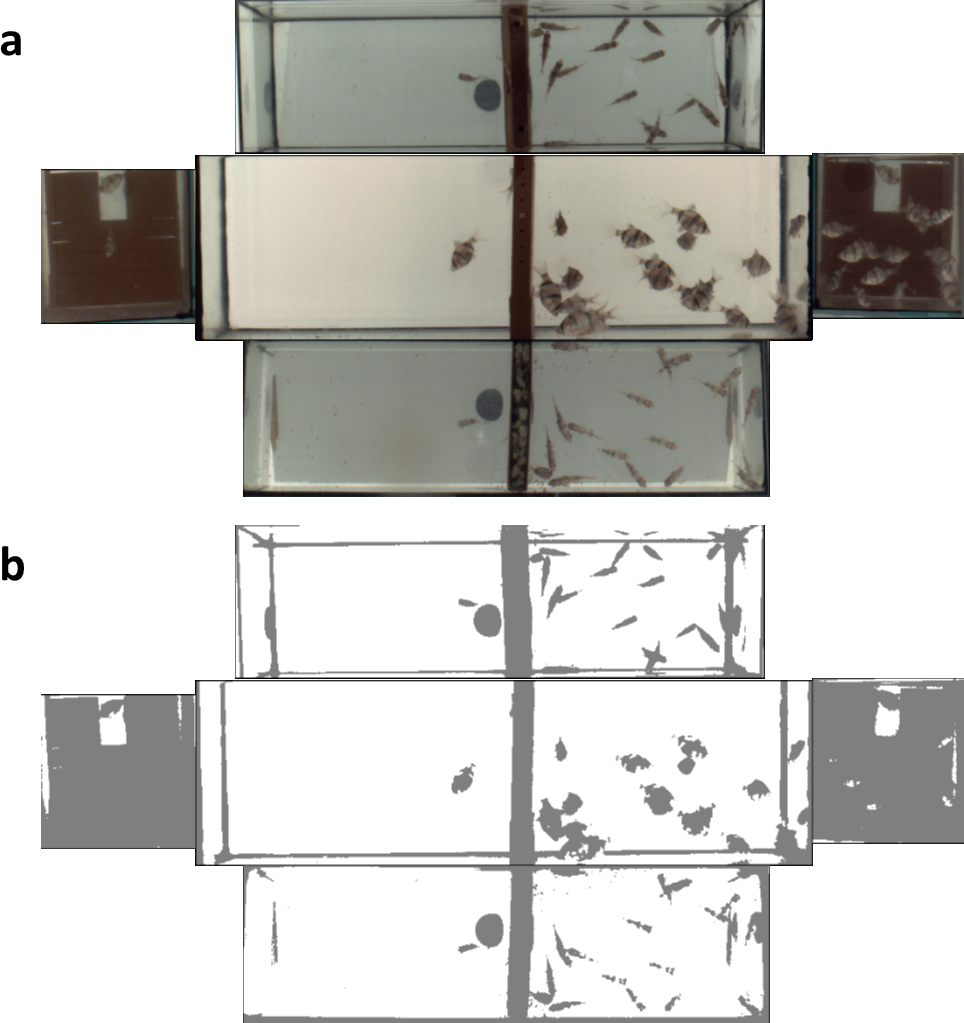}
\caption{Intensity thresholding of the mirrors in the corrected image. Colours of pixels were clustered into to groups using $k$-means++ algorithm into 2 groups (squared Euclidian measure). The uninformative black pixels were removed.}
\label{Fig6}
\end{figure}

The analysis was aimed to identify (1) the regions of the aquarium and of the mirrors (i.e., all regions were fish can occur) in the original image and (2) images of the fish in a raw file with the manually selected respective regions. As seen in Figs.~\ref{Fig4}--\ref{Fig6}, the method systematically failed in both tasks when applied to the uncorrected images and succeeded in corrected images. In task (1), the vignetting in the uncorrected image (Fig.~\ref{Fig4}a) joined the center of the image (aquarium) to the background (a wall of the illumination box). In task (2), in case of the uncorrected image, the repeatability of the $k$-means thresholding resulted in three different  binary images (Fig.~\ref{Fig5}) due to the trimodal shape (similar to those in Fig.~\ref{Fig2}a) of the intensity histograms. Providing the clustering into two groups, each from the three peaks is always assigned to a different group of pixels. Furthermore, unlike the corrected image (Fig.~\ref{Fig4}), in Fig.~\ref{Fig5}b, the vignetting prevented the proper detection of the background in the mirrors and, in Figs.~\ref{Fig5}c--d, the signal was overthresholded.

\section{Conclusions}
 
The calibration method described in this article utilizes experimental physico-chemical approach based on the measurement of light spectra reaching each camera (image) pixel separately. Unlike the methods based on calibration in color space, e.g.~\cite{Fang,Funt}, this approach makes this method objectively undoubted. Without any extensive technical treatments, the method is applicable to each digital camera. At the combination of different colour filters, light intensity and statistical analysis, it is able to estimate the whole light spectrum reaching each camera pixel.

The application of the calibration presented here showed that, for uncalibrated images, many computer analyses can fail. Since standard calibration methods utilize 8-bit images which are heavily distorted by a 12-bit to 8-bit compression~\cite{Stys2016}, we could not compare them with our method. 

We propose that, for scientific purposes and for computer-based image analyses, the images should be firstly corrected by our new method. The corrected images properly reflect incoming light. The calibration even gives fully comparable matrices of numbers of incoming photons which provides an ultimately correct dataset for comparison of outputs of any type of digital cameras. Such corrected images (raw camera files) can be processed by very simple algorithms which can decrease the computation costs significantly. But, if the data are simply wrong, the sophisticated algorithms can unavoidably fail as well.

\subsubsection*{Acknowledgments.} This work was supported by the Ministry of Education, Youth and Sports of the Czech Republic -- projects CENAKVA (No.\linebreak CZ.1.05/2.1.00/01.0024), CENAKVA II (No. LO1205 under the NPU I program), The CENAKVA Centre Development (No. CZ.1.05/2.1.00/19.0380) -- and from the European Regional Development Fund in frame of the project Kompetenzzentrum MechanoBiologie (ATCZ133) in the Interreg V-A Austria--Czech Republic programme.

\end{document}